\documentclass[journal]{vgtc}                     
\usepackage{graphicx}
\usepackage[linesnumbered,ruled,vlined]{algorithm2e}
\usepackage{amsmath}
\usepackage{amsfonts}
\usepackage{amssymb}
\usepackage{makeidx}
\usepackage{multirow}
\usepackage{cite}

\usepackage{algpseudocode}
\usepackage{lettrine}
\usepackage[switch]{lineno}

\onlineid{0}

\vgtccategory{Research}

\vgtcpapertype{please specify}

\title{
InterMamba: Efficient Human-Human Interaction Generation with Adaptive Spatio-Temporal Mamba
}

\author{%
  \authororcid{Zizhao Wu}{0000-0003-2103-5037},
  Yingying Sun, 
  Yiming Chen,
  Xiaoling Gu,
  Ruyu Liu
  and 
  Jiazhou Chen
}

\authorfooter{
  \item
  	Zizhao Wu is with the department of digital media technology, Hangzhou Dianzi University, Hangzhou 310018, China.
  	E-mail: wuzizhao@hdu.edu.cn
  \item
  	Yingying Sun is with the department of digital media technology, Hangzhou Dianzi University, Hangzhou 310018, China.
  	E-mail: yingyingsun@hdu.edu.cn
  \item
  	Yiming Chen is with the department of digital media technology, Hangzhou Dianzi University, Hangzhou 310018, China.
  	E-mail:; chenyiming@hdu.edu.cn
  \item
  	Xiaoling Gu is with the department of computer science, Hangzhou Dianzi University, Hangzhou 310018, China.
  	E-mail: guxl@hdu.edu.cn.
  \item
  	Ruyu Liu is with the school of information science and technology, Hangzhou Normal University, Hangzhou, 311121, China. Email: lry@hznu.edu.cn.
  \item
  	Jiazhou Chen is with the college of computer science and technology, Hangzhou
Zhejiang University of Technology, Hangzhou 310023, China. E-mail: 
cjz@zjut.edu.cn
}

\abstract{%
Human-human interaction generation has garnered significant attention in motion synthesis due to its vital role in understanding humans as social beings. However, existing methods typically rely on transformer-based architectures, which often face challenges related to scalability and efficiency. To address these issues, we propose a novel, efficient human-human interaction generation method based on the Mamba framework, designed to meet the demands of effectively capturing long-sequence dependencies while providing real-time feedback.
Specifically, we introduce an adaptive spatio-temporal Mamba framework that utilizes two parallel SSM branches with an adaptive mechanism to integrate the spatial and temporal features of motion sequences. To further enhance the model's ability to capture dependencies within individual motion sequences and the interactions between different individual sequences, we develop two key modules: the self adaptive spatio-temporal Mamba module and the cross adaptive spatio-temporal Mamba module, enabling efficient feature learning.
Extensive experiments demonstrate that our method achieves the state-of-the-art results on
both two interaction datasets with remarkable quality and efficiency. Compared to the baseline method InterGen, our approach not
only improves accuracy but also requires a minimal parameter size of just 66M—only 36\% of InterGen’s—while achieving an average
inference speed of 0.57 seconds, which is 46\% of InterGen’s execution time. 
}

\teaser{
  \centering
  \label{fig:teaser}
  \includegraphics[width=1\linewidth]{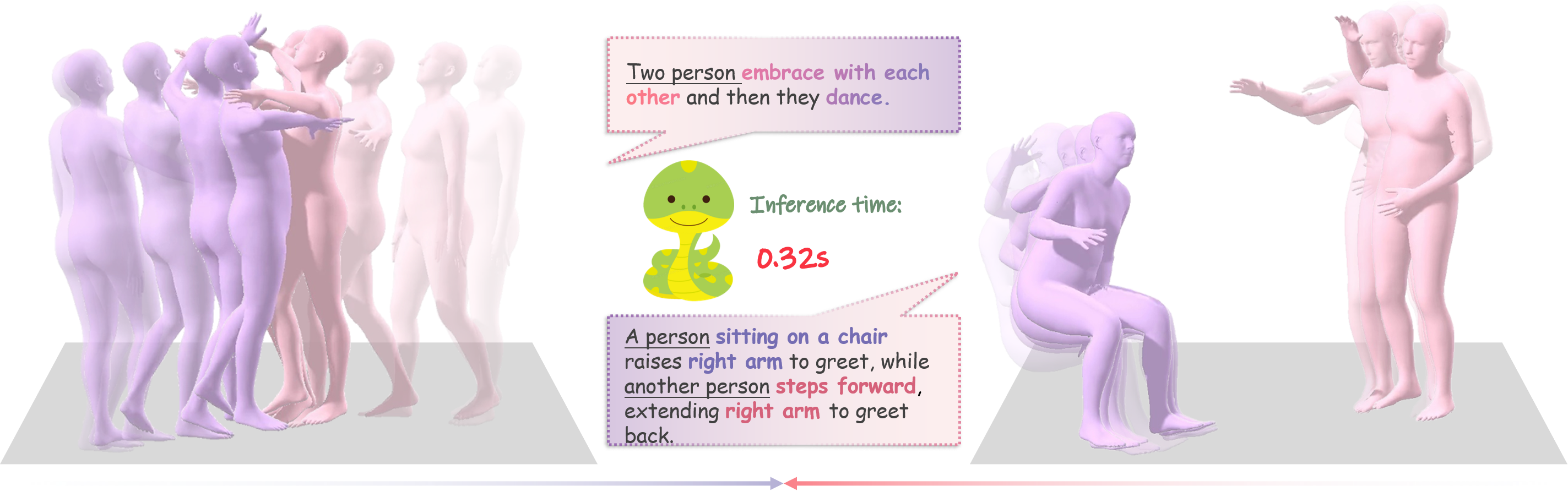}
  \caption{In this paper, we introduce an efficient human-to-human interaction generation method based on the Mamba framework, designed to achieve real-time, high-fidelity motion synthesis.}
}

\renewcommand{\manuscriptnotetxt}{}

\graphicspath{{figs/}{figures/}{pictures/}{images/}{./}} 

\usepackage{tabu}                      
\usepackage{booktabs}                  
\usepackage{lipsum}                    
\usepackage{mwe}                       

\usepackage{mathptmx}                  

\begin{document}

\firstsection{Introduction}

\maketitle

Human-Human interaction generation (HHI) is a crucial component in many graphics and vision applications, such as gaming, virtual simulation, and embodied AI. In recent years, with the rapid development of deep generative models \cite{XuSWSFDGYJYZ023, WuMZWZ24, OordVK17}, particularly diffusion models \cite{ho2020denoising}, significant progress has been achieved for single-person human motion generation in terms of controllability \cite{kim2023flame,wang2021scene}, diversity \cite{guo2022generating}, and quality \cite{zhang2023finemogen}. Building on these advancements, researchers have begun exploring more complex, interaction-centered human motion generation\cite{liang2024intergen, priorMDM, Ruiz-Ponce_2024_CVPR}, including Human-Human interaction generation.

Compared to single-human motion generation, human-human interaction generation is more challenging, as it requires simultaneously ensuring high-quality individual motion generation while accurately modeling interactions between individuals. To address these issues, significant efforts have been made, leading to notable progress. For example, InterGen\cite{liang2024intergen} introduces a diffusion-based transformer framework, where each individual's latent features are conditioned on those of others through cross-attention mechanisms that share weights, yielding promising results. ComMDM\cite{priorMDM} employs lightweight neural layers to bridge pre-trained single-person motion diffusion models, effectively enabling multi-person interaction modeling. InterMask \cite{javed2025intermask} proposes a masked VQVAE framework for the task in the discrete space by utilizing the VQVAE \cite{OordVK17} and Inter-M transformer.
However, we note that most existing methods are built upon the Transformer architecture. While Transformer-based approaches effectively model intricate contextual relationships, benefiting from their powerful attention mechanism, they also inherit the drawback of quadratic complexity, which limits their scalability and efficiency for long sequences.
Although various techniques such as windowing\cite{liu2021Swin}, sliding\cite{pan2023slide}, sparsification\cite{pietruszka2022sparsifyingtransformermodelstrainable}, hashing\cite{miao2024localitysensitivehashingbasedefficientpoint}, Ring Attention\cite{liu2023ringattentionblockwisetransformers}, and Flash Attention\cite{dao2023flashattention2}—or their combinations—have been introduced to mitigate the quadratic complexity of the Transformer, it remains a bottleneck in terms of scalability and efficiency.

Recently, Structured State Space Models (SSMs) have gained attention for their efficiency and effectiveness in handling long sequences. Mamba \cite{gu2021combiningrecurrentconvolutionalcontinuoustime,gu2023mamba}, building on this foundation, offers a promising solution for capturing long-range dependencies by introducing input selectivity and leveraging the scan method. Unlike transformers, which scale quadratically with sequence length, Mamba achieves linear or near-linear complexity while maintaining strong performance on extended sequences. This advantage has positioned it as a leading approach in continuous long-sequence analysis, delivering the state-of-the-art results in areas like natural language processing and computer vision. 
Therefore, it is natural to introduce Mamba into the human motion generation paradigm to solve the scalability problem of existing transformer-based methods. 

However, directly applying the existing Vanilla Mamba to our human-human interaction framework presents the following two challenges: Firstly, the vanilla Mamba module is specialized in 1D sequence modeling; how to extend it to process the spatio-temporal 3D motion sequences remains a challenge. The key lies in achieving an accurate fusion of spatio-temporal features while maximizing efficiency. Secondly, due to the complex interactions between human motions, another critical challenge is how to extend Mamba to model intricate multi-person interactions and communication effectively.

To tackle the above challenges, in this paper, we propose an Adaptive Spatio-Temporal Mamba (ASTM) framework for human-human interaction generation, which leverages SSM operations along temporal and spatial dimensions, with adaptive parameters to dynamically adjust their respective contributions, resulting in high scalability and efficiency in capturing the context of long-range sequences. 
To achieve precise and efficient modeling of spatiotemporal single-person motion sequences and complex human-to-human motion interactions, we propose a Self-ASTM module and a Cross-ASTM module, respectively. Specifically, the Self-ASTM captures long-range dependencies within individual characters by leveraging attention-based spatio-temporal Mamba, while the Cross-ASTM explicitly establishes connections between paired individuals and facilitates interaction-aware information exchange through the Local Interaction Information Aggregation (LIIA).

\begin{figure}[t!]
    \centering
    \includegraphics[width=1\linewidth]{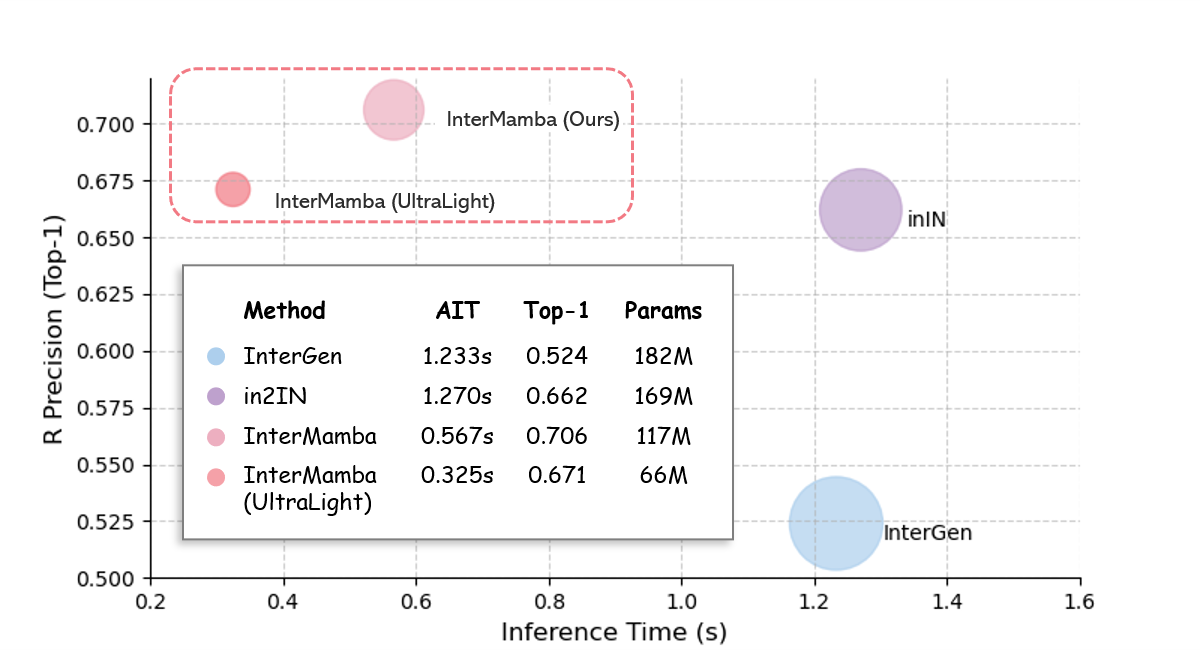}
    \caption{This figure presents a comparative analysis of different methods on the Interhuman dataset in terms of R-Precision (Top-1) and inference time, with the bubble size indicating the number of model parameters. From the figure, our InterMamba and InterMamba (UltraLight), which are highlighted in a red dashed box, achieve superior performance compared to other methods}
    \label{fig:top_list}
\end{figure}

We conduct our experiments on two famous benchmark datasets: InterHuman \cite{liang2024intergen} and InterX \cite{xu2024inter}, and the results have demonstrated that our approach achieves superior performance and lower computational footprint (as shown in Fig. \ref{fig:top_list}), setting new precedents for efficient, scalable human-human interaction generation. 

In summary, we make the following three contributions in this paper:
\begin{itemize}
    \item[$\bullet$] We propose InterMamba, a novel Mamba-based framework for human-to-human interaction generation. By leveraging adaptive spatio-temporal SSMs, our approach effectively captures long-range spatial and temporal dependencies, providing a highly efficient framework for modeling complex interactions.
    
    \item[$\bullet$] We explore the Self-ASTM and the Cross-ASTM modules, which are specifically designed to capture long-range dependencies within individual motion sequences and model the intricate interactions between individuals, respectively.
    
    \item[$\bullet$] Our method achieves the state-of-the-art results, demonstrating significant improvements in both generation quality and real-time efficiency.
\end{itemize}

\section{Related Work}
In this section, we briefly review relevant literature on single-human motion generation, human-human interaction generation, as well as SSMs.

\subsection{Single-Human Motion Generation.}
Single-human Motion Generation involves synthesizing single human movements based on various control conditions, including past motions \cite{martinez2017human,hernandez2019human, WuMZWZ24, PengMW23}, music \cite{li2021ai,tseng2023edge,aristidou2022rhythm}, action labels \cite{guo2020action2motion,petrovich2021action}, scenes \cite{cao2020long,wang2021scene}, and text \cite{kim2023flame, LiangBZRXY0YX24, pinyoanuntapong2024mmm,guo2024momask}. Among these conditions, text-driven motion generation has gained significant attention due to its ability to provide a more intuitive and accessible way to represent human intent. Methods like MotionCLIP \cite{tevet2022motionclip} integrate CLIP-based semantic knowledge into motion representations, while approaches such as TEMOS \cite{petrovich2022temos} and T2M \cite{guo2022generating} employ Transformer-based VAEs to generate motion from textual descriptions. Further advancements, including AttT2M \cite{ZhongHZX23}  and TM2D \cite{GongLCGJZMW23}, incorporate spatial-temporal body-part encoders into VQ-VAE to enhance the expressiveness of discrete motion representations.

Recently, diffusion models have emerged as a dominant paradigm for human motion generation due to their strong generative capabilities and diversity. Methods like Motion Diffusion Model (MDM) \cite{TevetRGSCB23}, MotionDiffuse \cite{zhang2022motiondiffuse}, and FLAME \cite{kim2023flame} adopt diffusion models with Transformer-based backbones to enable various conditioning strategies for coherent motion synthesis. Since these approaches operate directly on raw motion data, making them computationally expensive and susceptible to noise. To address this, MLD \cite{chen2023executing} introduces latent-space diffusion to improve motion quality, OmniControl \cite{xie2024omnicontrol} and FineMoGen \cite{zhang2023finemogen} enhance text-driven motion generation with more precise spatial-temporal control. Additionally, ReMoDiffuse \cite{ZhangGPCHLYL23} and Make-An-Animation \cite{AzadiSHPG23}  extend the generalization ability of diffusion models by incorporating large-scale in-the-wild data, such as human pose images and videos, into the training process.

Recent research highlights that human motions in real-world applications are inherently influenced by interactions with surrounding objects, other individuals, and the environment. These findings have led to growing interest in interaction-centered motion generation, encompassing tasks such as human-object interaction \cite{LiCMWPL24,abs-2312-06553}, human-human interaction \cite{priorMDM,liang2024intergen, Ruiz-Ponce_2024_CVPR}, and human-scene interaction \cite{JiangZLMWCLZ024, WangCJLZZL00024}. In this work, we focus specifically on the task of human-human interaction generation.

\subsection{Human-Human Interaction Generation.}
Compared to single-human motion generation, human-human interaction generation is more challenging, as it requires simultaneously ensuring high-quality individual motion generation while accurately modeling interactions between individuals. 
Many researchers have made valuable contributions to address this challenge. For example, building on the strong performance of MDM \cite{TevetRGSCB23}, ComMDM \cite{priorMDM} applies fine-tuning and minor modifications to adapt it for various motion-related tasks. Notably, ComMDM pioneers the task of human-human interaction motion generation by incorporating a communication transformer block between two instances of a pre-trained MDM \cite{TevetRGSCB23}.
InterGen \cite{liang2024intergen} makes a significant contribution to this area by introducing a dedicated human interaction dataset and proposing an interaction-aware diffusion model. This model leverages a novel weight-sharing mechanism with a global motion representation, achieving high-quality human interaction synthesis.
The in2IN \cite{Ruiz-Ponce_2024_CVPR} further advances the field by introducing a diffusion model that conditions motion generation not only on overall interaction descriptions but also on individual actions. This approach enhances both intra-person motion diversity and inter-person coordination, improving the realism and coherence of generated interactions.

Although the above-mentioned methods have achieved promising results, we observe that they are all built on the Transformer architecture, which suffers from quadratic time complexity, posing limitations in handling long motion sequences and computational efficiency. Therefore, this paper focuses on addressing the scalability and efficiency challenges of existing approaches by proposing a novel Mamba-based human-human interaction motion generation method.

\subsection{Selective State Space Models.}
Recently, State Space Models (SSMs) \cite{gu2021combiningrecurrentconvolutionalcontinuoustime, hasani2022liquid, gu2023mamba, fu2022hungry} have demonstrated remarkable performance in handling long-context tasks, offering both high-quality generation and efficient inference. However, the earlier SSMs struggled with effectively modeling discrete and information-dense data. To address this limitation, Mamba \cite{gu2023mamba} introduced a data-dependent SSM layer and a selection mechanism utilizing parallel scan (S6), significantly enhancing its applicability across various domains.

Since then, Mamba has sparked widespread interest in the deep learning community, leading to a surge of research in computer vision. SSMs are now proving their vast potential across diverse applications. For instance, Vim \cite{zhu2024vision} presents a novel vision backbone integrating bidirectional Mamba blocks for visual representation learning, achieving modeling capabilities comparable to ViT \cite{dosovitskiy2021imageworth16x16words} while reducing computational complexity to subquadratic time with linear memory requirements. VMamba \cite{liu2024vmambavisualstatespace} introduces 2D Selective Scan (SS2D), a four-way scanning mechanism designed for spatial domain processing, ensuring that each image patch captures rich contextual information. VideoMamba \cite{park2024videomambaspatiotemporalselectivestate} extends these concepts to video understanding, combining the strengths of convolution and attention within a pure SSM framework, demonstrating efficiency and effectiveness in both short-term and long-term video tasks. Similarly, PointMamba \cite{liu2024point} adapts SSMs for 3D vision applications by enabling global modeling with linear complexity, making it a compelling choice for point cloud analysis.
In the motion domain, MotionMamba \cite{zhang2024motion} introduces a hierarchical temporal Mamba block alongside a bidirectional spatial Mamba block to facilitate real-time and high-quality single-human motion generation. 

In this work, we introduce Mamba into the human-human interaction motion generation paradigm for the first time. We show that, compared to other methods, our approach achieves significant improvements in both generation quality and efficiency.

\section{Preliminaries}
\label{preliminies}
\subsection{Diffusion model} Following DDPM \cite{ho2020denoising} and InterGen \cite{liang2024intergen}, our model adopts a diffusion framework with forward and reverse processes for motion generation. The forward process adds time-dependent noise to the motion data \( p_0(x) \), transforming it into \( p_t(x) \) over \( t \) steps until it approximates \( \mathcal{N}(0, \mathbf{I}) \). The reverse process removes the noise to reconstruct \( p_0(x) \) via a generative model parameterized by \( \theta \). According to \cite{ho2020denoising}, we simplify the multi-step diffusion process into a single step, expressed as:
{\baselineskip=0.5\baselineskip 
\begin{equation}
    \begin{split}
    & q(x_t \mid x_0) = \mathcal{N}\left(\sqrt{\bar{\alpha}_t} x_0, (1 - \bar{\alpha}_t)\mathbf{I}\right)
    \end{split},
\end{equation}
}
where $\bar{\alpha}_t$ lies within the range $(0, 1)$ and follows a monotonically decreasing schedule. As $t$ approaches infinity, $\bar{\alpha}_t$ converges to 0, causing $d_t$ to approximate a sample from the standard normal distribution. The reverse process aims to learn $f_\theta$ to iteratively reconstruct the motion, generating $\hat{x}_0$ conditioned on the input text $c$. Instead of predicting the noise, we directly estimate $\hat{x}_0$, following the approach in \cite{zhang2022motiondiffuse}. The training objective is defined as:
{\baselineskip=0.9\baselineskip 
\begin{equation}
    \mathcal{L}_{\text{t}} = \mathbb{E}_{x_0, t}\left[\|x_0 - f_\theta(x_t, t, c)\|_2^2\right].
\end{equation}
}

\begin{figure}[t!]
    \centering
    \includegraphics[width=1\linewidth]{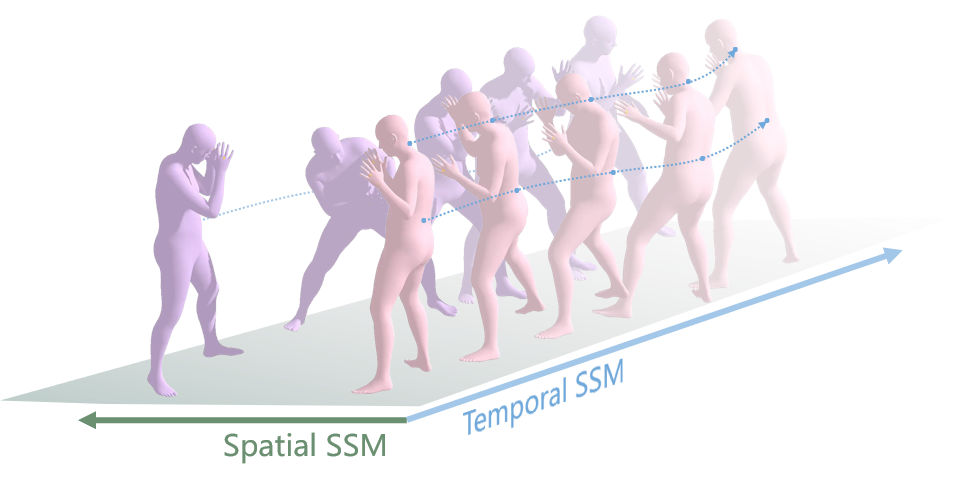}
    \caption{The spatial and temporal scanning process, where the spatial SSM captures intra-frame joint relationships and the temporal SSM models inter-frame relationships.}
    \label{fig:STscan}
\end{figure}

\subsection{Mamba Model}  
Selective state space models (SSMs) which have emerged as a promising tool for sequence modeling with efficient computation and data storage. Inspired by a particular continuous system, traditional SSMs maps a 1-dimensional function or sequence $x(t)$ to $y(t)$ through an implicit latent state $h(t)$.
\begin{equation}
\label{eq:SSM1}
    \begin{split}
    & h'(t) = \textbf{A}h(t)+\textbf{B}x(t)  \\
    & y(t) = \textbf{C}h(t) ,
    \end{split}
\end{equation}
\noindent where $\textbf{A} \in \mathbb{R}^{N \times N}$ denotes the evolution parameter, and  $\textbf{B} \in \mathbb{R}^{N \times 1}$, $\textbf{C} \in \mathbb{R}^{1 \times N}$ denotes the projection parameters. However, these parameters are time-invariant matrices and vectors, which constrains capacities on modeling discrete and information-dense data. To resolve this problem, Mamba proposes a novel selection mechanism that parameterizes the SSM parameters based on the input data to affect interactions along the sequence. The rewritten architecture formulation is defined as below:
\begin{equation}
\label{eq:SSM2}
    \begin{split}
    & h_t = \bar{\textbf{A}}_t h_{t-1}+\bar{\textbf{B}}_t x_t \\
    & y_t = \textbf{C}_t h_t ,
    \end{split}
\end{equation}
\noindent where $\bar{\textbf{A}}_t$ , $\bar{\textbf{B}}_t$ , and $\textbf{C}_t$ are dynamically updating matrixs and vectors by the time step, meaning that the model has changed from time-invariant to time-varying. As the discrete versions of the continuous system, Mamba \cite{gu2023mamba} transform the continuous parameters $\textbf{A}$, $\textbf{B}$ to discrete parameters $\bar{\textbf{A}}_t$, $\bar{\textbf{B}}_t$ with a timescale parameter $\Delta$. The discretization rule, zero-order hold (ZOH), is defined as follows:

\begin{equation}
\label{eq:SSM3}
    \begin{split}
    & \bar{\textbf{A}} = exp(\Delta \bold{A})\\
    & \bar{\textbf{B}} = (\Delta \textbf{A})^{-1}(exp(\Delta\textbf{A})-\bold{I})\cdot \Delta\textbf{B},
    \end{split}
\end{equation}

\noindent where $(\Delta \textbf{A})^{-1}$ denotes the inverse of matrix $\Delta\textbf{A}$, $\bold{I}$ denotes the identity matrix. Additionally, Mamba imply a Selective Scan Mechanism (SSMs) as its core SSM operator. Thus, mamba makes these parameters change depending on the input which can be formulated as:
\begin{equation}
\label{eq:SSM4}
    \begin{split}
    & \textbf{B} = Linear_N(x)\\
    & \textbf{C} = Linear_N(x)\\
    & \Delta = \tau_{\Delta}(P+LayerNorm(Linear_1(x)) , \\
    \end{split}
\end{equation}
where $P$ is a learnable parameter, $ \tau_{\Delta}$ is  SoftPlus function, and $Linear_d(\cdot)$ is a parameterized projection to dimension $d$. The parameters $\textbf{B}\in \mathbb{R}^{(B, L, N)}$, $\textbf{C} \in \mathbb{R}^{(B, L, N)}$, and $\Delta \in \mathbb{R}^{(B, L, D)}$ are directly derived from the input data $x\in \mathbb{R}^{(B, L, D)}$. 

\begin{figure*}[t]
    \includegraphics[width=0.95\textwidth]{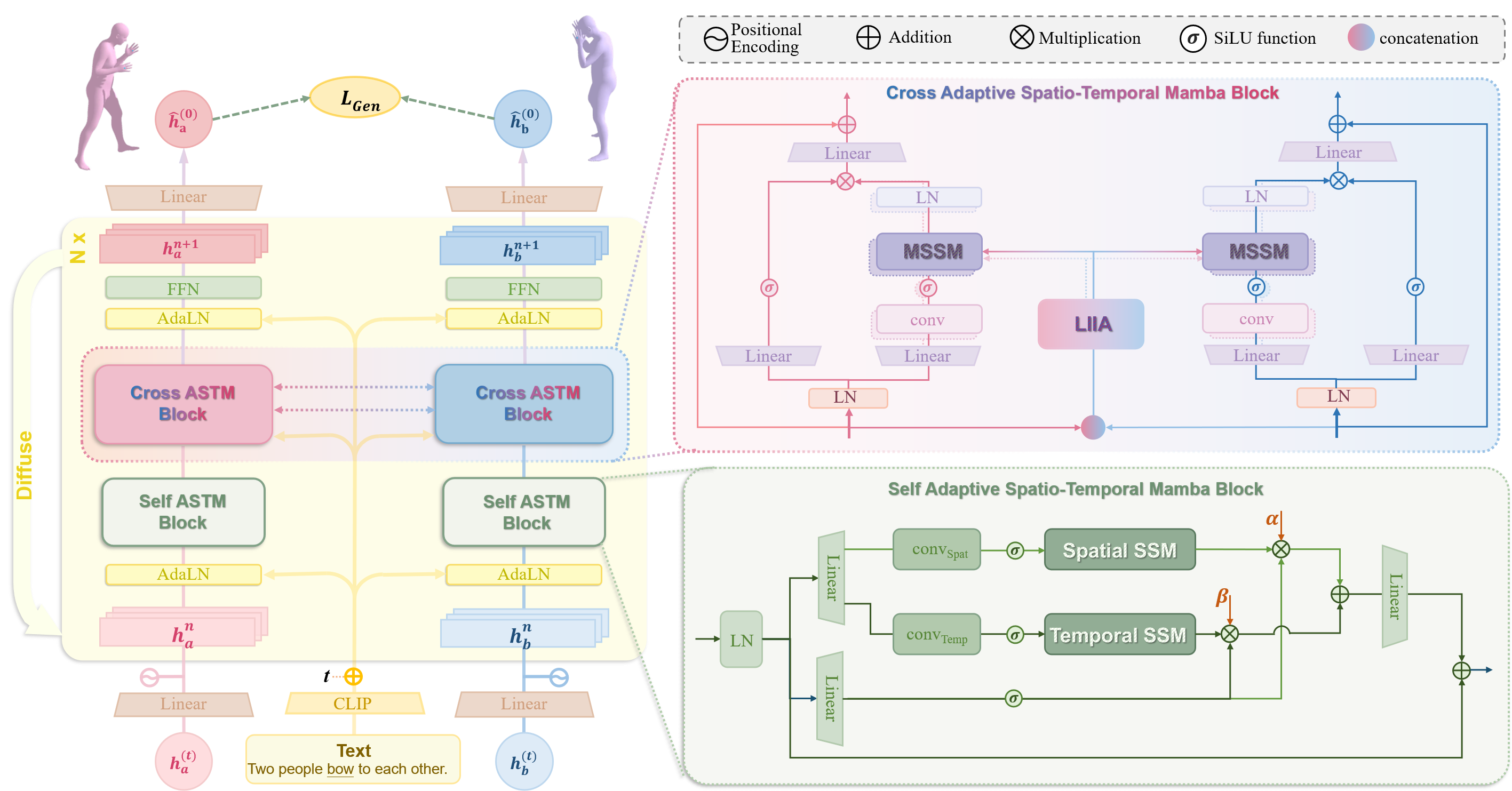}
    \caption{
    \textbf{The framework of InterMamba.} The key components include Self Adaptive Spatio-Temporal Mamba module (Self-ASTM), Cross Adaptive Spatio-Temporal Mamba module (Cross-ASTM), and Local Interaction Information Aggregation module (LIIA). }
    \label{fig:framework}
\end{figure*}

\section{Method}
Our objective is to generate a high-quality human-human interaction motion sequence that aligns
with the action described in the text prompt while maintaining high efficiency. In our implementation, we use $\{x_p\} = \{x_p^i\}_{i=1}^n, p \in \{a,b\}$ to denote the outcome motion sequences, where $n$ is the frame number of the sequence. For each pose $x_p^i, p \in \{a,b\}$, we define it as: ${x}_{p}^i  = [j^g, v^g, r^l, c^f], p \in \{a,b\}$, where $j^g \in \mathbb{R}^{3J} $ denotes the global joint positions, $v^g \in \mathbb{R}^{3J}$ represents the global velocities, $r^l \in \mathbb{R}^{6J}$ refers to 6D representation of local rotations and $c^f \in \mathbb{R}^{4}$ corresponds to the binary foot-ground contact features. 
Figure \ref{fig:framework} illustrates the pipeline of our method, which mainly consists of Adaptive Spatio-Temporal Mamba modules, Self-Role Mamba Blocks, Cross-Role Mamba Blocks and LIIA interaction modules. 
Specifically, given a text prompt, our method first employs the CLIP encoder to achieve the textual features embedding. Then the textural features are combined with the initial hidden states \( h_a^{(t)} \) and \( h_b^{(t)} \) separately and are fed into our framework. Within the framework, our method employs the Self-ASTM Blocks for modeling the long-range motion contexts of individuals and Cross-ASTM Blocks for modeling the intricate interaction contexts between individuals. To further enhance the representation, we exploit the LIIA module that aggregates human-human interaction representations. After $N$ iterations, our method ultimately generates an efficient and high-quality human-human interaction sequence that aligns with the semantics of the text input.

In the following, we begin with the introduction of the backbone of our framework in Sec. \ref{sec:AST}.  In Sec. \ref{sec:SRM}, we detail the Self-ASTM module. Next, we present LLIA module in Sec. \ref{sec:LIIC}. In Sec. \ref{sec:CRM}, we introduce the Cross-ASTM module.  Finally, in Sec. \ref{sec:objfun} we provide the objective function of our method. 

\subsection{Adaptive Spatio-Temporal Mamba}
\label{sec:AST}

As described in \Cref{preliminies}, the Mamba model is inherently well-suited for sequential modeling due to its selection architectural design. However, its feature extraction is primarily focused on the temporal dimension, lacking explicit modeling of spatial dynamics.  Consequently, Mamba exhibits performance bottlenecks in such spatio-temporal entangled scenarios. This limitation becomes apparent when addressing complex tasks such as 3D human interaction generation.
To overcome this challenge, we propose a novel parallel spatio-temporal modeling framework named Adaptive Spatio-Temporal Mamba (ASTM). ASTM decomposes the modeling process into two complementary branches: a Spatial State Space Model (Spatial SSM) and a Temporal State Space Model (Temporal SSM).
Specifically, as illustrated in \Cref{fig:STscan}, Spatial SSM focuses on capturing intra-frame spatial dependencies, effectively extracting structural information such as the interactions between skeletal joints within each frame. In contrast, the Temporal SSM is dedicated to modeling inter-frame dynamics, capturing long-range motion trajectories and temporal causality.
To be specific,  given the input feature tensor $h \in \mathbb{R}^{(B, L, D)}$, where $B$ is the batch size, $L$ is the sequence length, and $D$ is the feature dimension, the temporal SSM processes the input as follows:
\begin{equation}
\label{spatial ssm}
    h_t = {LayerNorm}({SSM_{temp}}({Conv_{temp}}({Linear}(h)))),
\end{equation}
where $\text{Linear}(\cdot)$ projects input to a latent space, $Conv_{spat}(\cdot)$ extracts local spatial patterns, $SSM_{temp}(\cdot)$ is $SSM$ based module which captures long-range temporal dependencies, and $\text{LayerNorm}(\cdot)$ is layer normalization which improves training stability and convergence.
Similar to the temporal branch, the spatial SSM adopts the same module structure but operates along the spatial dimension. The spatial output can be computed by:
\begin{equation}
\label{temporal ssm}
h_s = {LayerNorm}({SSM_{spat}}({Conv_{spat}}({Linear}(h')))),
\end{equation}
where all modules are identical in structure, the input $h' \in \mathbb{R}^{(B, D, L)}$  is derived by transposing the input tensor to swap the sequence and feature dimensions.
To adaptively integrate the learned spatial and temporal representations, we introduce a learnable fusion mechanism. Instead of naïvely concatenating or averaging features, we allow the model to dynamically reweight the contributions from $h_s$ and $h_t$. The final fused representation is computed as:
\begin{equation}
z=w_\alpha h_t + w_\beta h_s,
\end{equation}
where $w_\alpha$ and $w_\beta$  are jointly optimized during training.

Benefiting from this decoupled modeling and adaptive fusion approach, ASTM significantly enhances the model’s capacity to represent the spatio-temporal characteristics of complex motion, leading to improved structural coherence and temporal consistency in human interaction generation tasks.

\begin{figure*}[htbp]
\label{ssm and cssm}
        \centering
    \begin{subfigure}[htb]{1\columnwidth}
  	\centering
  	\includegraphics[width=1 \columnwidth]{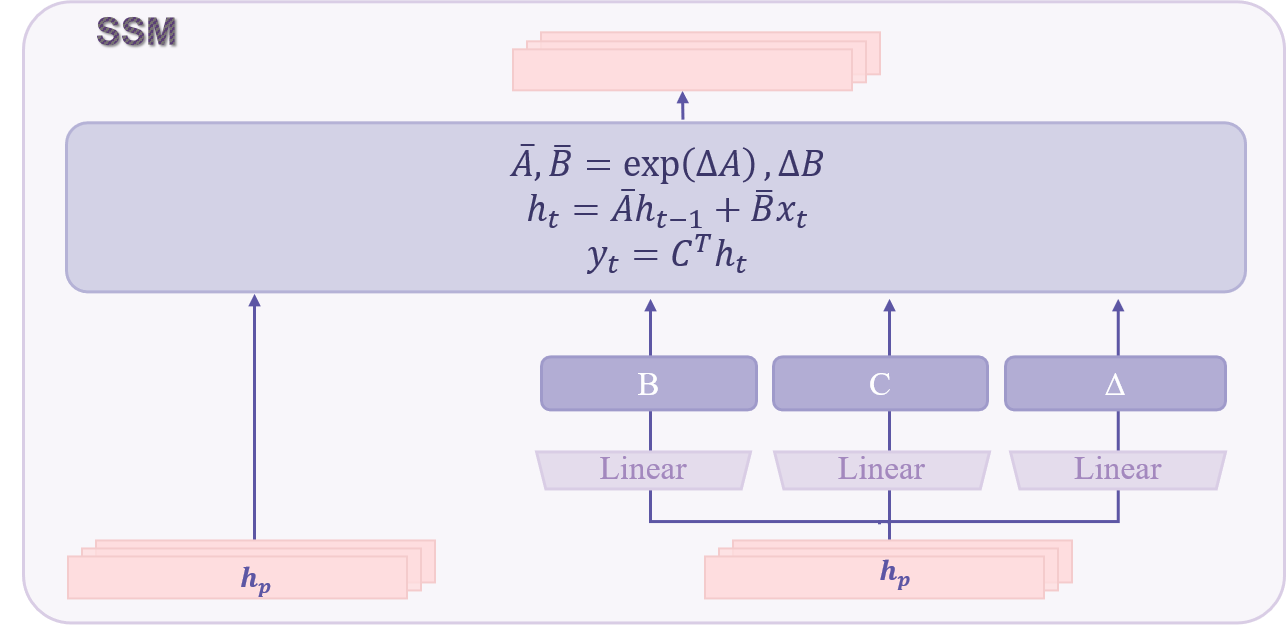}
  	\caption{Schematic Diagram of the State Space Model (SSM)}
  	\label{SSM_self}
  \end{subfigure}%
  \hfill%
  \hfill%
     \begin{subfigure}[tb]{1\columnwidth}
    	\centering
  	\includegraphics[width=1 \columnwidth]{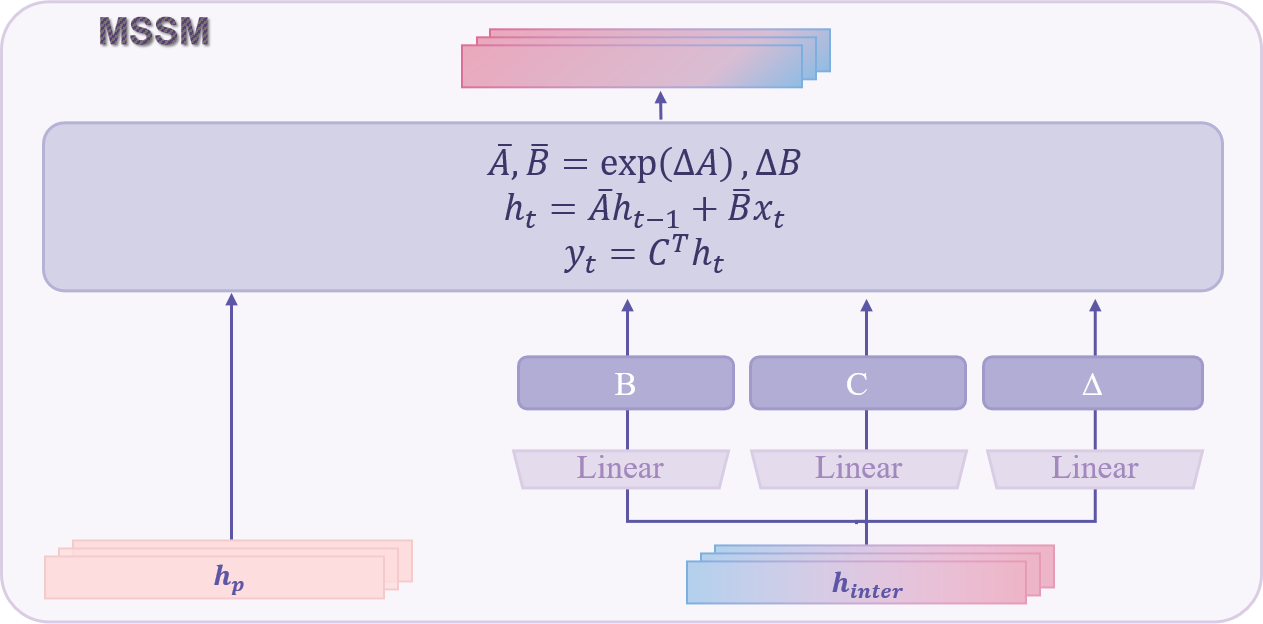}
  	\caption{Schematic Diagram of the Mixed State Space Model (Mix-SSM)}
  	\label{CSSM}
     \end{subfigure}
    \subfigsCaption{(a) This figure illustrates the core module of the ASTM in \Cref{sec:AST} —— \textbf{State Space Model (SSM)}. Given an input sequence $h_p$, the input first undergoes linear transformations through \Cref{eq:SSM3} to calculate the parameters $B$, $C$, and $\Delta$. Then, the final output can be calculated through \Cref{eq:SSM2} and \Cref{eq:SSM3}. (b) This figure illustrates the core module of the $\mathrm{ASTM}_{\text{cross}}$ in \Cref{sec:CRM} —— \textbf{Mix State Space Model (MSSM)}. This module, based on the State Space Model (SSM), enables each individual to perceive the mutual interaction features dynamically, leading to improved modeling of contextual dependencies. Given $h_p$ and $h_{\text{inter}}$, which is calculated from \Cref{sec:LIIC}, the parameters $B$, $C$, and $\Delta$ can be calculated through $x_{\text{inter}}$, and $x_t$ is the input sequence of the CSSM. Then, the final output can be calculated through \Cref{eq:SSM2} and \Cref{eq:SSM3}. The details of Mix SSM are provided in \Cref{sec:CRM}.}  
  \label{fig:SSM&CSSM}
\end{figure*}

\subsection{ Self-ASTM block}
\label{sec:SRM}
To enable the model to effectively learn the motion features of each individual in human-human interaction, we designed the self Adaptive Spatio-Temporal Mamba Block (Self-ASTM) based on the ASTM, as illustrated in \Cref{fig:framework}. This module processes the motion of each individual separately as input and leverages ASTM to extract the most salient spatio-temporal features for each character. Specifically, given the motion feature of a single individual $h_p^{(t)},p\in {\{a,b\}}$ as input, the output feature can be calculated as:
\begin{equation}
    \label{eq:5}
    \begin{split}
        &\bar h_{\text{p}} = LayerNorm(h_p), \\
        & {\hat{h}_p} = ASTM(\bar{h}_{\text{p}}),\\
        & q = \sigma(Linear(\bar{h}_{\text{p}})),\\
        & \widetilde{h}_p = \hat{h}_p \star q,\\
        & h_{p}^S = h_p + Linear(\widetilde{h}_p),
    \end{split}
\end{equation}
where $LayerNorm(\cdot)$ is the layer normalization, $Linear(\cdot)$ is the linear layer for feature mapping, $\sigma$ is the activation function and the SiLU function is used in our experiment. $ASTM(\cdot)$ can be calculated from \Cref{sec:AST}.

\subsection{Local Interaction Information Aggregation}
\label{sec:LIIC}

\begin{figure}[t!]
    \centering
    \includegraphics[width=0.95\linewidth]{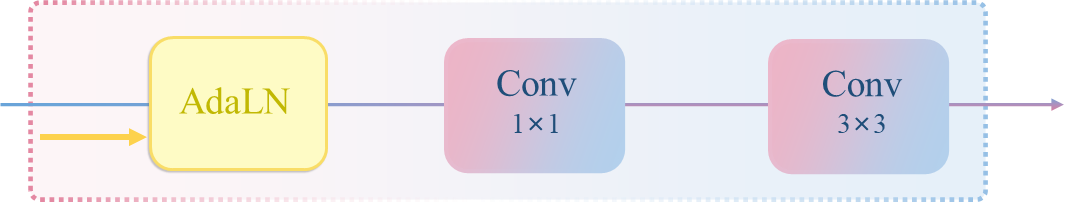}
    \caption{The structure of Local Interaction Information Aggregation (LIIA) consists of AdaLN for adaptive normalization, followed by 1×1 and 3×3 convolutions for feature refinement and local interaction modeling.}
    \label{fig:LIIC}
\end{figure}

Information in dyadic interactions is typically highly abstract and densely entangled. However, existing methods often rely on raw individual features and utilize mutual attention mechanisms to model feature-level relationships. Those approaches are prone to introducing noise and struggle to align deeper semantic representations and interactive information.
To address this issue, we investigate Local Interaction Information Aggregation(LIIA) module, as illustrated in \Cref{fig:LIIC}, which consists of two convolution layers and a residual connection. Given the motion feature of a single individual $h_p,p\in {\{a,b\}}$ and text feature embeddings $cond_t$, the output joint representation
 $h_{inter}$ can be formulated as:
\begin{equation}
\label{eq:LIIC}
\begin{split}
    & h_{ab}=concat(h_a^S,h_b^S), \\
    & h_{ab}=AdaLN(h_{ab},cond_t),\\ 
    & h_{inter}=conv_{3 \times 3}(conv_{1 \times 1}(h_{ab})). \\
\end{split}
\end{equation}
where $conv_{k \times k}$ denotes the convolution with kernel size $k$, $AdaLN(\cdot)$ is the adaptive layer normalization to reduce the impact of noise features on the input and optimize the training process. We note that this strategy not only facilitates the modeling of local feature interactions but also ensures a more comprehensive and robust integration of contextual information.

\begin{table*}[htbp]
  \caption{%
  	\textbf{The quantitative comparisons on the InterHuman\cite{liang2024intergen} test set.} We run all the evaluations 20 times. ± indicates a 95\% confidence interval. ↑ (↓) indicates that a higher (lower) result corresponds to better performance, and → means the closer to ground-truth the better. \textbf{Bold} indicates the best result, and \underline{underline} refers to the second best.%
  }
  \label{tab:interhuman}
  \fontsize{7}{10}\selectfont
  \centering%
  \setlength{\aboverulesep}{1.2pt}
  \setlength{\belowrulesep}{1.2pt}
  \renewcommand{\arraystretch}{0.5} 
  \resizebox{\textwidth}{!}{ 
  \begin{tabu}{%
  	  l%
  	  	*{7}{c}%
  	}
  	\toprule
    \multirow{2}{*}{$Methods$} & \multicolumn{3}{c}{$R Precision\uparrow$} & \multirow{2}{*}{$FID\downarrow$} & \multirow{2}{*}{$MM Dist \downarrow$} & \multirow{2}{*}{$Diversity\rightarrow$} & \multirow{2}{*}{$MModality\uparrow$} \\ 
    \cmidrule{2-4} %
    & Top1 & Top2 & Top3 &  \\ \midrule
  	Real & $0.452^{\pm .008}$ & $0.610^{\pm .009}$ & $0.701^{\pm .008}$ & $0.273^{\pm .003}$ & $3.755^{\pm .007}$ & $7.948^{\pm .064}$ & - \\
  	TEMOS \cite{petrovich2022temos} & $0.224^{\pm .010}$ & $0.316^{\pm .013}$ & $0.450^{\pm .018}$ & $17.375^{\pm .043}$ & $5.342^{\pm .015}$ & $6.939^{\pm .071}$ & $0.535^{\pm .014}$ \\
  	T2M \cite{guo2022generating} & $0.238^{\pm .012}$ & $0.325^{\pm .010}$ & $0.464^{\pm .014}$ & $13.769^{\pm .072}$ & $4.731^{\pm .013}$ & $7.046^{\pm .022}$ & $1.387^{\pm .076}$ \\
  	MDM \cite{TevetRGSCB23} & $0.153^{\pm .012}$ & $0.260^{\pm .009}$ & $0.339^{\pm .012}$ & $9.167^{\pm .056}$ & $6.125^{\pm .018}$ & $7.602^{\pm .045}$ & $2.355^{\pm .080}$ \\
  	ComMDM* \cite{priorMDM} & $0.067^{\pm .013}$ & $0.125^{\pm .018}$ & $0.184^{\pm .015}$ & $38.643^{\pm .098}$ & $13.211^{\pm .013}$ & $3.520^{\pm .058}$ & $0.217^{\pm .018}$ \\
  	ComMDM \cite{priorMDM} & $0.223^{\pm .009}$ & $0.334^{\pm .008}$ & $0.466^{\pm .011}$ & $7.069^{\pm .054}$ & $5.212^{\pm .021}$ & $7.244^{\pm .038}$ & $1.822^{\pm .052}$ \\
  	RIG \cite{tanaka2023RIG} & $0.285^{\pm .010}$ & $0.409^{\pm .014}$ & $0.521^{\pm .013}$ & $6.775^{\pm .069}$ & $4.876^{\pm .018}$ & $7.311^{\pm .043}$ & $\mathbf{2.096^{\pm .065}}$ \\
  	InterGen \cite{liang2024intergen} & $0.371^{\pm .010}$ & $0.515^{\pm .012}$ & $0.624^{\pm .010}$ & $5.918^{\pm .079}$ & $5.108^{\pm .012}$ & $7.387^{\pm .029}$ & $2.141^{\pm .063}$ \\
    in2IN \cite{Ruiz-Ponce_2024_CVPR} & $0.425^{\pm .008}$ & $0.576^{\pm .006}$ & $0.662^{\pm .009}$ & $\mathbf{5.535^{\pm .120}}$ & $3.803^{\pm .002}$ & $7.983^{\pm .047}$ & $1.215^{\pm .023}$ \\
    \midrule
  	\textbf{InterMamba (Ours)} & $\mathbf{0.475^{\pm .005}}$ & $\mathbf{0.625^{\pm .006}}$ & $\mathbf{0.706^{\pm .004}}$ & $\underline{5.945^{\pm .007}}$ & $\mathbf{3.785^{\pm .003}}$ & $\mathbf{7.963^{\pm .033}}$ & $0.993^{\pm .026}$ \\
     \textbf{InterMamba (light)} & $\underline{0.443^{\pm .005}}$ & $\underline{0.592^{\pm .005}}$ & $\underline{0.671^{\pm .004}}$ & $8.082^{\pm .095}$ & $\underline{3.801^{\pm .004}}$ & $\underline{7.991^{\pm .033}}$ & $0.965^{\pm .020}$ \\
  	\bottomrule
  \end{tabu}%
  }
\end{table*}

\begin{table*}[htbp]
  \caption{%
  	\textbf{The quantitative comparisons on the Inter-X \cite{xu2024inter} test set.} We run all the evaluations 20 times. ± indicates a 95\% confidence interval. \textbf{Bold} indicates the best result, and \underline{underline} refers to the second best.%
  }
  \label{tab:interX}
  \fontsize{7}{10}\selectfont
  \centering%
    \setlength{\aboverulesep}{1.2pt}
  \setlength{\belowrulesep}{1.2pt}
    \renewcommand{\arraystretch}{0.5} 
  \resizebox{\textwidth}{!}{ 
  \begin{tabu}{%
  	  l%
  	  	*{7}{c}%
  	}
  	\toprule
    \multirow{2}{*}{$Methods$} & \multicolumn{3}{c}{$R Precision\uparrow$} & \multirow{2}{*}{$FID\downarrow$} & \multirow{2}{*}{$MM Dist \downarrow$} & \multirow{2}{*}{$Diversity\rightarrow$} & \multirow{2}{*}{$MModality\uparrow$} \\ 
    \cmidrule{2-4} %
    & Top1 & Top2 & Top3 &  \\ \midrule
  	Real & $0.429^{\pm .004}$ & $0.626^{\pm .003}$ & $0.736^{\pm .003}$ & $0.002^{\pm .0002}$ & $3.536^{\pm .013}$ & $9.734^{\pm .078}$ & - \\
  	TEMOS \cite{petrovich2022temos} & $0.092^{\pm .003}$ & $0.171^{\pm .003}$ & $0.238^{\pm .002}$ & $29.258^{\pm .0694}$ & $6.867^{\pm .013}$ & $4.738^{\pm .078}$ & $0.672^{\pm .041}$ \\
  	T2M \cite{guo2022generating} & $0.184^{\pm .010}$ & $0.298^{\pm .006}$ & $0.396^{\pm .005}$ & $5.481^{\pm .3820}$ & $9.576^{\pm .006}$ & $5.771^{\pm .151}$ & $2.761^{\pm .042}$ \\
        T2M* \cite{guo2022generating} & $0.325^{\pm .004}$ & $0.487^{\pm .005}$ & $0.593^{\pm .005}$ & $4.506^{\pm .020}$ & $8.5347^{\pm .055}$ & $5.771^{\pm .151}$ & $2.761^{\pm .042}$ \\
  	MDM \cite{TevetRGSCB23}& $0.203^{\pm .009}$ & $0.329^{\pm .007}$ & $0.426^{\pm .005}$ & $23.701^{\pm .0569}$ & $9.548^{\pm .014}$ & $5.856^{\pm .077}$ & $\mathbf{3.490^{\pm .061}}$ \\
  	MDM(GRU) \cite{TevetRGSCB23} & $0.179^{\pm .006}$ & $0.299^{\pm .005}$ & $0.387^{\pm .007}$ & $32.617^{\pm .1221}$ & $9.557^{\pm .019}$ & $7.003^{\pm .134}$ & $3.430^{\pm .035}$ \\
  	ComMDM \cite{priorMDM}& $0.090^{\pm .002}$ & $0.165^{\pm .004}$ & $0.236^{\pm .003}$ & $29.266^{\pm .0668}$ & $6.870^{\pm .017}$ & $4.734^{\pm .067}$ & $0.771^{\pm .053}$ \\
  	InterGen \cite{liang2024intergen} & $0.207^{\pm .004}$ & $0.335^{\pm .005}$ & $0.429^{\pm .005}$ & $5.207^{\pm .2160}$ & $9.580^{\pm .018}$ & $7.788^{\pm .208}$ & $3.686^{\pm .052}$ \\ 
    \midrule
  	\textbf{InterMamba (Ours)} & $\mathbf{0.4573^{\pm .004}}$ & $\mathbf{0.642^{\pm .005}}$ & $\mathbf{0.742^{\pm .004}}$ & $\mathbf{0.517^{\pm .026}}$ & $\mathbf{3.594^{\pm .018}}$ & $\mathbf{9.194^{\pm .066}}$ & $2.593^{\pm .080}$ \\
  \bottomrule
  \end{tabu}%
  }
\end{table*}

\subsection{Cross-ASTM Block}
\label{sec:CRM}

While the Self-ASTM efficiently models individual motion features, it overlooks the interaction details between individuals, which are crucial for human-human interaction generation task. To address this problem,  inspired by \cite{lin2024mtmamba}, we introduce a novel cross Adaptive Spatio-Temporal Mamba block (Cross-ASTM) which extends the Self-ASTM to explicitly model cross-individual interaction relationships.  As illustrated in \Cref{fig:framework}, the overall architecture of the cross-ASTM mirrors that of the Self-ASTM, with one key difference: the original $ASTM$ module is replaced but the newly designed $ASTM_{cross}$ module. Specifically, given the inputs $h_p^S \in \mathbb{R}^{(B, L, D)}$ and $h_{inter} \in \mathbb{R}^{(B, L, D)}$, obtained through the Self-STM and LIIA, the output of the Cross-ASTM block can be calculated as:
\begin{equation}
    \label{eq:CASTM}
    \begin{split}
        &\bar h_{p} = \text{LayerNorm}(h_p^S), \\
        & \hat{h}_p = ASTM_{cross}(\bar{h}_{p},h_{inter}),\\
        & q = \hat{\sigma}(Linear(\bar{h}_{p})),\\
        & \widetilde{h}_p = \hat{h}_p \star q,\\
        & h_{\text{out}} = h_p^S + \text{Linear}(\widetilde{h}_p),
    \end{split}
\end{equation}
where $LayerNorm(\cdot)$ is the layer normalization, $Linear(\cdot)$ is the linear layer for feature mapping, $\hat{\sigma}$ is the activation function. Within this module, the $ASTM_{cross}$ function is designed to extract spatio-temporal features by jointly considering both individual representations and their interactions. It is based on the Mix State Space Model (MSSM) architecture and extends it by introducing two parallel branches dedicated to spatial and temporal modeling, respectively. The internal computation of this mechanism is formally defined as follows: 
\begin{equation}
\label{CSSM_ST}
\begin{split}
    & c_s = {LayerNorm}({MSSM_{temp}}({Conv_{temp}}({Linear}(h_p^S),h_{inter}))), \\
    & c_t = {LayerNorm}({MSSM_{soat}}({Conv_{spat}}({Linear}({h^S_p}'),h'_{inter}))) \\
    &\hat{h}_p = \alpha_c*c_s + \beta_c*c_t , \\
\end{split}
\end{equation}
where the ${h_p^S}' \in \mathbb{R}^{(B, D,L)}$ and $h_{inter}' \in \mathbb{R}^{(B, D,L)}$ are obtained by swapping the last two dimensions of $h_p^S$ and $h_{\text{inter}}$, respectively, so that the inputs are aligned along the spatial dimension. The parameters $\alpha_c$ and $\beta_c$ are learnable scalars.

The modules $MSSM_{\text{spat}}$ and $MSSM_{\text{temp}}$ are instantiated from the base MSSM, which extends the standard State Space Model (SSM) by integrating interaction information. Unlike conventional SSMs that operate solely on individual feature sequences, MSSM simultaneously processes both individual features and interaction features for enhanced feature fusion. Specifically, the interaction features $h_{\text{inter}}$ are used to generate the parameters $(B, C, \Delta)$, while the individual features $h_p$ serve as the query input, as defined in \Cref{eq:SSM4}. The output is computed using \Cref{eq:SSM2} and \Cref{eq:SSM3}.
This design enables the model to better capture and emphasize interaction-relevant information. The architecture of MSSM is illustrated in \Cref{fig:SSM&CSSM}.

\subsection{The Objective Function}
\label{sec:objfun}
We adopt the same loss functions as those used in InterGen \cite{liang2024intergen}, which encompass several loss functions, and can be formulated as:

\begin{equation}
\label{eq:loss}
\begin{aligned}
\mathcal{L}_{total}= & \mathcal{L}_{diff} + \lambda_{vel}\mathcal{L}_{vel}+\lambda_{foot}\mathcal{L}_{foot}+\lambda_{BL}\mathcal{L}_{BL} \\
 & +\lambda_{DM}\mathcal{L}_{DM}+\lambda_{RO}\mathcal{L}_{RO},
\end{aligned}\end{equation}
where $\mathcal{L}_{diff}$ refers to the diffusion loss, $\mathcal{L}_{vel}$ denotes the joint velocity loss, $\mathcal{L}_{foot}$ refers to the foot contact loss, $\mathcal{L}_{BL}$ represents the bone length loss, $\mathcal{L}_{DM}$ corresponds to the masked joint Distance Map (DM) loss and $\mathcal{L}_{RO}$ signifies the relative orientation loss.  All loss terms are weighted by hyper-parameters $\lambda_{vel}$, $\lambda_{foot}$, $\lambda_{BL}$, $\lambda_{DM}$ and $\lambda_{RO}$, that balance their influence on the final result.


\section{Experiments}
In this section, we first outline the details of our experimental setup in Sec. \ref{sec:eSetup}, followed by both quantitative and qualitative evaluations to assess the effectiveness of our method in Sec. \ref{sec:Comparison with State-of-the-art Methods}. In Sec. \ref{sec:ablation study}, we present a comprehensive ablation study. Additionally, we provide a supplementary video for further qualitative analysis.

\subsection{Experiment Setup}
\label{sec:eSetup}
\noindent\textbf{Datasets}. 
We conduct our experiments on two publicly available benchmark datasets: InterHuman \cite{liang2024intergen} and InterX \cite{xu2024inter}.
InterHuman \cite{liang2024intergen} is the first human-human interaction dataset with natural language annotations. It contains 6,022 motion sequences spanning various categories of human actions, annotated with 16,756 unique descriptions composed of 5,656 distinct words, with a total duration of 6.56 hours.
InterX \cite{xu2024inter} adopts the SMPL-X representation, which includes 55 joints covering the face, body, and hands. The dataset consists of 11,000 interaction sequences, accompanied by more than 34,000 fine-grained textual annotations at the body-part level, totaling over 8.1 million frames.
\\
\noindent\textbf{Implementation Details}. 

Our framework consists of two stacked ASTM modules (with an ultra-light variant containing only one ASTM module). Each ASTM module employs a state size of 16. For text encoding, we utilize the frozen CLIP-ViT-L/14 model. The diffusion process is trained with 1,000 timesteps, while inference employs the DDIM sampling strategy with 50 timesteps and $\eta = 0$. We adopt the cosine noise schedule \cite{nichol2021improved} and classifier-free guidance \cite{ho2022classifier}, setting random CLIP embeddings to zero with a probability of 10\% during training and using a guidance coefficient of 3.5 during sampling. The model is optimized using the AdamW optimizer with a learning rate of $10^{-4}$ and a weight decay of $2\times10^{-5}$. All models are trained for 2,000 epochs with a batch size of 50 on two NVIDIA RTX 4090 GPUs. And the inference is conducted on a single NVIDIA RTX 4090 GPU.

\begin{figure*}
    \centering
    \includegraphics[width=1\linewidth]{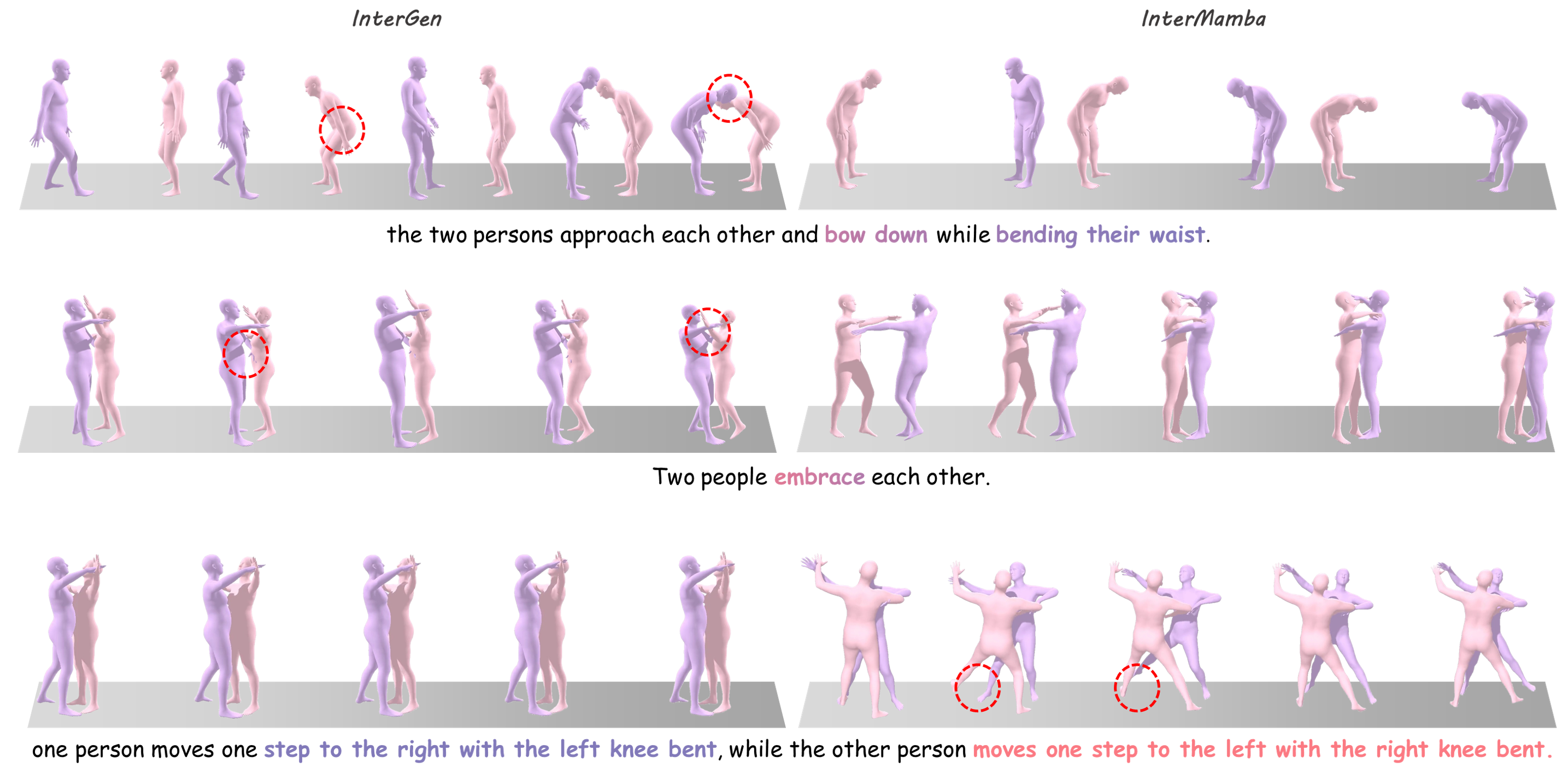}
    \caption{
    Quantitative comparison of state-of-the-art methods on the text-driven 3D human-human interaction generation task, accompanied by qualitative visualizations using three representative prompts. The untrustworthy actions are highlighted within the red circle.}
    \label{fig:qual comp}
\end{figure*}

\noindent\textbf{Evaluation Metrics}. 
We adopt the same evaluation metrics as InterGen \cite{liang2024intergen}  to assess our model's performance, which include:
(1) \textbf{Frechet Inception Distance (FID):} measures the discrepancy in latent distributions between the generated and real datasets.  
(2) \textbf{R-Precision:} evaluates text-motion alignment by determining the probability of the correct text appearing in the top-$k$ results ($k = 1, 2, 3$) after ranking. (3) \textbf{Diversity:} quantifies the variation in generated motions within the dataset. (4) \textbf{Multimodality (MModality):} assesses the diversity of motions generated for the same text input. (5) \textbf{Multi-modal Distance (MM Dist):} calculates the distance between motion features and their corresponding text features.

\subsection{Comparison with Baselines}
\label{sec:Comparison with State-of-the-art Methods}
\subsubsection{Qualitative Comparison}
To ensure a comprehensive comparison, we evaluate our approach against several baseline methods, including TEMOS \cite{petrovich2022temos}, T2M \cite{guo2022generating}, MDM \cite{TevetRGSCB23}, ComMDM \cite{priorMDM}, InterGen \cite{liang2024intergen}, RIG \cite{tanaka2023RIG} and in2IN \cite{Ruiz-Ponce_2024_CVPR}. 
Among these, InterGen \cite{liang2024intergen} represents a milestone in the field as it introduces the first human-human interaction dataset, while in2IN \cite{Ruiz-Ponce_2024_CVPR} is the most recent advancement. Therefore, we mainly discuss the performance of our method against these two approaches.

Tables \ref{tab:interhuman} and \ref{tab:interX} present the quantitative comparison of our InterMamba with various human-human interaction generation methods. Each evaluation is conducted 20 times, except for MModality, which is run 5 times, and we report the averaged results along with a 95\% confidence interval.
The results demonstrate that InterMamba consistently outperforms previous approaches, achieving the highest R-Precision across all settings (0.475/0.625/0.706 on InterHuman and 0.4573/0.642/0.742 on Inter-X), indicating superior text-motion alignment. Additionally, it attains the lowest MM Dist (3.785 on InterHuman and 3.594 on Inter-X), ensuring improved structural coherence and smoother motion transitions.
Furthermore, InterMamba exhibits the highest Diversity scores (7.963 on InterHuman and 9.194 on Inter-X), reflecting its strong capacity to generate a broad range of human interactions. While its FID scores (5.945 and 0.517, respectively) are slightly higher than a few baselines, this trade-off enhances motion quality and semantic consistency.
Additionally, InterMamba achieves a competitive MModality score (0.993 on InterHuman and 2.593 on Inter-X), demonstrating its ability to generate diverse and interaction-aware motion patterns. These results further validate that InterMamba effectively balances accuracy, coherence, and diversity, making it as the state-of-the-art solution for text-driven human-human interaction motion generation.

\subsubsection{Quantitative Comparison} 
In Fig. \ref{fig:qual comp}, we qualitatively compare our method with InterGen \cite{liang2024intergen}, both trained on the InterHuman dataset using identical text prompts.
For the prompt: \emph{The two persons approach each other and bow down while bending their waist}, InterGen exhibits unnatural limb positioning, rigid motion, and inconsistent spacing, lacking proper arm-torso contact. InterMamba, however, ensures fluid movement, synchronized waist bending, and realistic posture alignment. We attribute this to the design of ASTM, which selects different scanning directions to preserve spatial coherence.
For the prompt: \emph{Two people embrace each other}, InterGen struggles with spatial coordination, leading to misaligned hand positions and an unnatural embrace. In comparison, InterMamba produces a well-executed, natural interaction with realistic body contact. We attribute this to the LIIA module, which effectively captures fine-grained interaction details.
For the prompt: \textit{One person moves one step to the right with the left knee bent, while the other moves one step to the left with the right knee bent}, InterGen struggles to maintain correct foot placement and knee bending, due to its weaker ability to capture fine-grained textual details. In contrast, InterMamba, leveraging AdaLN and Mamba’s strong contextual information extraction capabilities, accurately reflects the prompt, ensuring synchronized stepping and spatial coherence. Besides, InterMamba exhibits superior interaction naturalness and spatial awareness, producing smooth transitions and physically plausible inter-person contact.

Overall, these qualitative results highlight that InterMamba significantly surpasses InterGen in generating semantically accurate, structurally coherent, and interaction-consistent human motions, establishing it as a more effective solution for text-driven human-human interaction generation.

\subsubsection{Complexity Analysis}
To assess the efficiency of our approach, we conduct a complexity analysis comparing our method against several existing approaches. The results, summarized in Tab. \ref{tab:complexity}, include inference time, model parameters, and FLOPs (floating point operations per second).
From the table, InterGen exhibits the highest computational cost, requiring 1.233s for inference with 182M parameters and 68.33G FLOPs, making it computationally expensive. in2IN has a slightly higher inference time (1.260s) but with 160M parameters, indicating similar computational demands as InterGen, though FLOPs data is unavailable.
In contrast, our InterMamba significantly reduces computational complexity, achieving an inference time of 0.567s—nearly 2.2$\times$ faster than InterGen—while maintaining a more efficient model size (117M parameters, 33.53G FLOPs). Furthermore, InterMamba (UltraLight) optimizes efficiency even further, achieving the lowest inference time (0.325s) with just 66M parameters and 14.05G FLOPs, making it the most lightweight and computationally efficient solution.
Compared to InterGen and in2IN, InterMamba runs more than twice as fast with significantly lower FLOPs, demonstrating its scalability and efficiency for real-time applications.

\begin{table}[t!]
  \caption{
  This table compares the computational complexity of our method with that of other approaches, based on three indicators: average inference time, parameters, and FLOPs. As shown in the table, our InterMamba achieves the highest overall efficiency across all metrics.
  }
  \label{tab:complexity}
  \centering%
  \renewcommand{\arraystretch}{0.9} 
  \setlength{\aboverulesep}{1.5pt}
  \setlength{\belowrulesep}{1.5pt}
  \setlength{\heavyrulewidth}{0.1em}
  \setlength{\lightrulewidth}{0.05em} 
  \resizebox{\columnwidth}{!}{ 
    \begin{tabu}{l*{3}{c}}
      \toprule
      Methods & Inference time(s) & Params (M) & FLOPs (G) \\  
      \midrule
      InterGen \cite{liang2024intergen}         &  1.233  &  182  & 68.33  \\ 
      in2IN \cite{xu2024inter}                  &  1.260  &  160  & 33.72  \\
      \midrule
      Ours                                      &  0.567  &  117  & 33.53  \\ 
      Ours (UltraLight)                         &  \textbf{0.325}  & \textbf{66}  & \textbf{14.05} \\
      \bottomrule
    \end{tabu}%
  }
\end{table}

\subsection{Ablation study}
\label{sec:ablation study}
\begin{figure}[t!]
    \centering
    \includegraphics[width=0.9\linewidth]{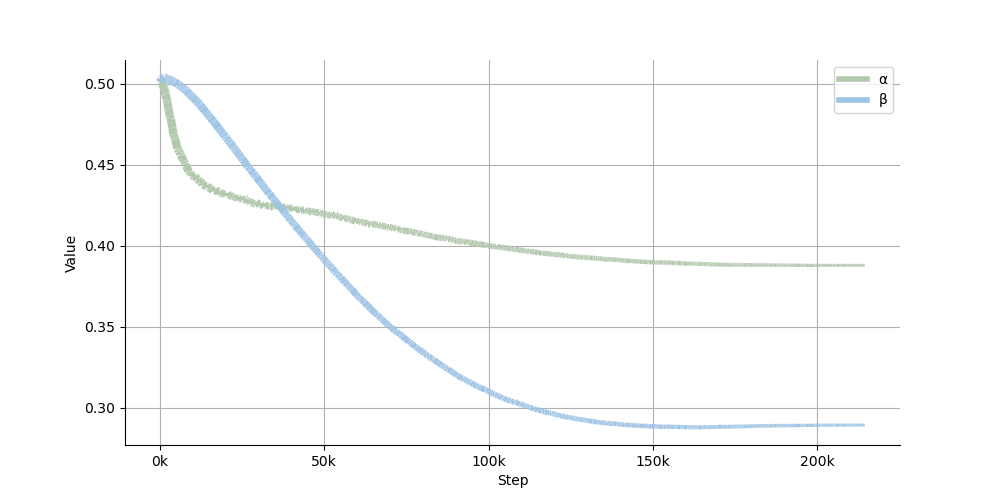}
    \caption{As the training progresses, the variation process of the adaptive parameters $\alpha$ and $\beta$. }
    \label{fig:LIIC_param}
\end{figure}

To assess the effectiveness of each module, we conduct comprehensive ablation study to quantify the significance of each module and explore their interactions within the overall framework. First, we present the importance of the adaptive parameters during training in \Cref{fig:LIIC_param}. Next, we examine the effectiveness of the three key components in our InterMamba framework by incrementally adding the proposed modules to the baseline InterGen \cite{liang2024intergen}, with the results shown in \Cref{tab:ASTSSM eff}.

\noindent \textbf{The Effect of Adaptive parameters.} The adaptive Spatio-Temporal SSM is the core component of both the Self-ASTM and the Cross-ASTM blocks. In this study, we use InterGen as the baseline for our method and investigate the effectiveness of the adaptive parameter mechanism. \Cref{fig:LIIC_param} illustrates the relationship between the parameters and the training steps of our approach.
From \Cref{fig:LIIC_param}, we observe that as the number of training steps increases, the parameter $\beta$ decreases significantly, indicating a stronger adaptation over time. Meanwhile, the parameter $\alpha$ remains relatively stable, suggesting that it continues to play a crucial role in the model. This result highlights that the adaptive mechanism enables the model to dynamically adjust the weighting of spatial and temporal components, which may enhance learning stability and guide the model toward convergence.
Table \ref{tab:ASTSSM eff} evaluates the model's performance under three different settings: without temporal SSMs, without adaptive parameters, and with adaptive spatio-temporal SSM. In the case without temporal SSMs, our method achieves a lower R-Precision score (0.679) but a better FID (5.539), indicating structurally coherent yet temporally inconsistent motion. In the second case, where adaptive parameters are removed, our model fails to converge, highlighting their critical role in ensuring stable training.
With the inclusion of adaptive spatio-temporal Mamba, our method achieves the highest R-Precision (0.705), demonstrating superior text-motion alignment, although there is a slight increase in FID (5.945) due to the added complexity of modeling long-term dependencies. These results confirm that adaptive parameterization is essential for convergence, while temporal SSMs improve motion consistency with only a minor trade-off in FID.

Overall, the Adaptive Spatio-Temporal mamba proves to be crucial for stability, temporal coherence, and accurate text-motion mapping, making it a fundamental component for generating realistic and high-fidelity human-human interaction generation.

\begin{table}[t!]
  \caption{%
  	The effectiveness of adaptive Spatio-Temporal SSM on the InterHuman dataset. \textbf{Bold} font highlights the best performance in each category.
  }
  \label{tab:ASTSSM eff}
  \centering%
  \renewcommand{\arraystretch}{1} 
  \fontsize{6.5}{9}\selectfont
  \setlength{\aboverulesep}{0.5pt}
  \setlength{\belowrulesep}{0.5pt}
  \setlength{\heavyrulewidth}{0.1em}
  \setlength{\lightrulewidth}{0.05em} 
  \resizebox{\columnwidth}{!}{ 
  \begin{tabu}{%
  	  l%
  	  	*{2}{c}%
  	}
  	\toprule
        Method & $FID\downarrow$ & $R-precision\uparrow$ \\
        \midrule
         w/o Temporal Mamba &$\mathbf{5.539^{\pm .102}}$ &  $0.679^{\pm .0.003}$ \\
         w/o adaptive param.&$6.775^{\pm .002}$ & $0.625^{\pm .0.012}$ \\
         w/ adaptive param.& $5.945^{\pm .007}$ &$\mathbf{0.705^{\pm .004}}$  \\
  	\bottomrule
  \end{tabu}%
  }
\end{table}

\noindent \textbf{Effect of Self-ASTM Block.}  
To demonstrate the superiority of the self-ASTM module, we compare its standalone performance against other configurations. The first row in Tab. \ref{tab:module effectiveness} shows that using only the Self-ASTM results in an R-Precision of \textbf{0.371} and an FID of \textbf{5.918}, indicating that SRM establishes a solid baseline for individual motion modeling. However, without inter-human interaction understanding, the model's text-motion alignment is suboptimal, highlighting the need for additional components to capture complex interactions.

\noindent \textbf{Effect of Cross-ASTM Block.}  
Adding the cross-ASTM block further improves the model's R-Precision to \textbf{0.409}, but significantly increases the FID to \textbf{8.524}. This trade-off highlights that while Cross-ASTM enhances inter-character motion coordination and interaction, it introduces challenges in maintaining consistent and coherent motions across the characters. As a result, the quality of motion generation degrades, leading to visual inconsistencies and a reduction in overall motion fluidity.

\noindent \textbf{Effect of Local Interaction Information Aggregation.}  
Further integrating the Local Interaction Information Aggregation (LIIA) alongside Self-ASTM and Cross-ASTM significantly boosts the R-Precision to \textbf{0.705}, the highest among all configurations, while maintaining a competitive FID of \textbf{5.945}. These results confirm that LIIA plays a crucial role in refining inter-human interactions and enhancing semantic consistency, effectively balancing high motion quality with strong text alignment.

\begin{table}[t!]
  \caption{%
  	\textbf{The effectiveness of each module in InterMamba on the InterHuman dataset is presented.} The Self-ASTM refers to the Self-ASTM Block in \Cref{sec:SRM}, Cross-ASTM denotes the Cross-ASTM Block in \Cref{sec:CRM}, and LIIA stands for the Local Interaction Information Aggregation module in \Cref{sec:LIIC}. Bold font highlights the best performance in each category.
  }
  \label{tab:module_effectiveness}
  \centering%
  \renewcommand{\arraystretch}{0.9} 
  \fontsize{9.5}{13}\selectfont
  \setlength{\aboverulesep}{1.5pt}
  \setlength{\belowrulesep}{1.5pt}
  \setlength{\heavyrulewidth}{0.1em}
  \setlength{\lightrulewidth}{0.05em} 
  \resizebox{1\columnwidth}{!}{%
  \begin{tabu}{*{3}{l}|*{3}{c}}
  	\toprule
  	Self-ASTM & Cross-ASTM & LIIA & R Precision (Top 1) $\uparrow$ & $FID\downarrow$ & Params (M) \\
    \midrule
    &    &    & $0.371^{\pm .010}$ & $\mathbf{5.918^{\pm .079}}$ & 182\\ 
    \checkmark & & &  $0.409^{\pm .012}$ & $8.524^{\pm .060}$ & 180 \\ 
    \checkmark & \checkmark & & $0.605^{\pm .002}$ & $13.184^{\pm .012}$ & 117 \\ 
    \checkmark & \checkmark & \checkmark & $\mathbf{0.705^{\pm .004}}$ & $5.945^{\pm .007}$ & $\mathbf{117}$\\ 
  	\bottomrule
  \end{tabu}%
  }
\end{table}

\section{Conclution}
In this paper, we introduce InterMamba, an efficient and high-quality framework for text-driven human-human interaction generation. Our approach effectively tackles two major challenges in the field: efficiently capturing individual motion features and interaction details from text while ensuring real-time inference speed. To achieve this, we propose an Adaptive Spatio-Temporal Mamba framework (ASTM) to dynamically extract and integrate spatial and temporal features. Building on this foundation, we develop two core modules: Self-ASTM, which captures long-range dependencies for each character using AST, and Cross-ASTM, which explicitly models human-human interactions with the assistance of Local Interaction Information Aggregation (LIIA). These modules enable InterMamba to generate realistic human-human interaction efficiently. Our extensive experiments demonstrate that InterMamba sets a new benchmark for real-time, high-quality human-human interaction generation, laying the groundwork for future advancements in text-driven Human-Human interaction generation.
\section{Limitations}
Despite InterMamba’s advancements, several limitations remain.
First, although inference speed is improved by integrating Mamba, the method still relies on diffusion-based generation, which involves multiple iterations. Reducing these steps speeds up inference but may affect motion quality. Future work could explore alternative models—like flow-based or hybrid approaches—to improve both speed and quality.
Second, while Mamba helps with long-range dependencies, the generated motions may still lack detail in subtle interactions, emotions, or physical realism. Enhancing contact modeling and biomechanical constraints is vital for more expressive results.
Finally, practical deployment needs improvement. Real-world applications like avatars and games require better user control, real-time adaptation, and multi-task optimization. Future work should focus on adaptive strategies like user feedback and multi-agent collaboration to bridge the gap between algorithms and real-world use.

\bibliographystyle{abbrv-doi-hyperref}

\bibliography{template}

\end{document}